\documentclass[sigplan, screen, nonacm]{acmart}

\usepackage{caption}
\usepackage{subcaption}
\usepackage{hyperref}
\usepackage{setspace}
\usepackage[ruled,linesnumbered]{algorithm2e}
\usepackage{multirow}
\usepackage{booktabs}
\usepackage{enumitem}
\usepackage{balance}

\AtBeginDocument{%
  }

\begin{document}

\title[TinySeg: Model Optimizing Framework for Image Segmentation on Tiny Embedded Systems]{TinySeg: Model Optimizing Framework for\\Image Segmentation on Tiny Embedded Systems}

\author{Byungchul Chae}
\orcid{0009-0006-8578-3228}
\affiliation{%
  \institution{Kyung Hee University}
  \city{Yongin}
  \country{South Korea}
}
\email{chaechae7@khu.ac.kr}

\author{Jiae Kim}
\orcid{0009-0005-0531-2884}
\affiliation{%
  \institution{Kyung Hee University}
  \city{Yongin}
  \country{South Korea}
}
\email{theyummy@khu.ac.kr}

\author{Seonyeong Heo}
\orcid{0000-0003-0359-1953}
\affiliation{%
  \institution{Kyung Hee University}
  \city{Yongin}
  \country{South Korea}
}
\email{seonyeong.heo@khu.ac.kr}


\begin{abstract}

Image segmentation is one of the major computer vision tasks, which is applicable in a variety of domains, such as autonomous navigation of an unmanned aerial vehicle. However, image segmentation cannot easily materialize on tiny embedded systems because image segmentation models generally have high peak memory usage due to their architectural characteristics. This work finds that image segmentation models unnecessarily require large memory space with an existing tiny machine learning framework. That is, the existing framework cannot effectively manage the memory space for the image segmentation models.

This work proposes TinySeg, a new model optimizing framework that enables memory-efficient image segmentation for tiny embedded systems. TinySeg analyzes the lifetimes of tensors in the target model and identifies long-living tensors. Then, TinySeg optimizes the memory usage of the target model mainly with two methods: (i) tensor spilling into local or remote storage and (ii) fused fetching of spilled tensors. This work implements TinySeg on top of the existing tiny machine learning framework and demonstrates that TinySeg can reduce the peak memory usage of an image segmentation model by 39.3\% for tiny embedded systems.

\end{abstract}



\maketitle

\section{Introduction}
\label{sec:intro}

\begin{figure}[t]
    \centering
    \vspace{20pt}
    \includegraphics[width=\hsize]{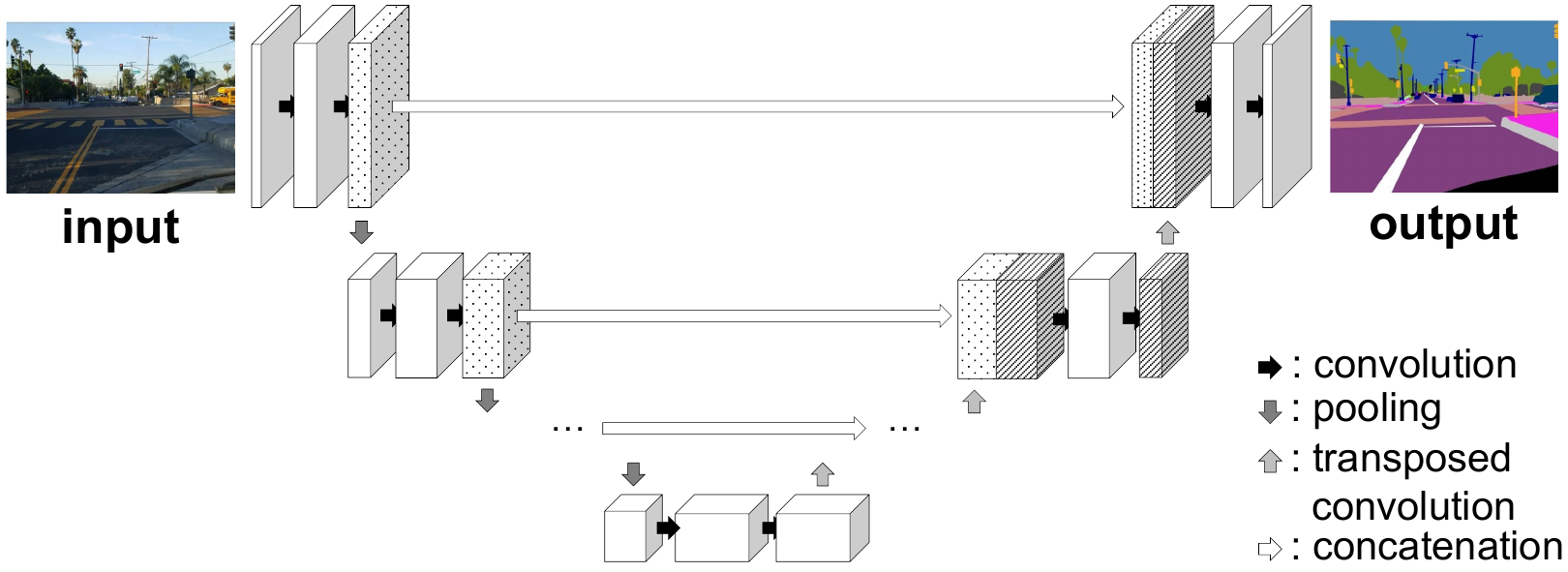}
    \caption{U-Net Network Architecture~\cite{ronneberger:2015:miccai}. Sample Images are Taken from the Cityscapes Dataset~\cite{cordts:2016:cvpr}.}
    \label{fig:unet}
\end{figure}

Image segmentation is one of the major computer vision tasks, which is being used in various domains, including autonomous navigation of an unmanned aerial vehicle~\cite{minaee:2021:tpami}. Image segmentation is to classify each pixel of the input image in an object category. For example, in autonomous driving, image segmentation can be used to detect cars, pedestrians, roads, and traffic signs~\cite{cordts:2016:cvpr}. Since image segmentation provides more detailed information about objects than object detection, image segmentation has become an attractive but challenging task for computer vision researchers.

Although image segmentation can be helpful for many applications, image segmentation cannot easily materialize on tiny embedded systems because of its high peak memory usage. Popular image segmentation networks~\cite{ronneberger:2015:miccai, milletari:2016:3dv, cao:2023:eccv} have an \textit{hourglass}-like architecture where the early computation results are concatenated with the later results. Then, the early results have to be stored in memory until the later results become available, which is the primary reason why image segmentation models generally have high peak memory usage. Since tiny embedded systems have limited memory of less than 1 megabyte, image segmentation models can hardly run on the systems, limiting the potential of intelligent embedded systems.

To demonstrate that image segmentation models unnecessarily take up large memory space, this work analyzes the memory consumption of image segmentation models on top of an existing tiny machine learning framework, TensorFlow Lite for Microcontrollers (TFLM)~\cite{tflm:web}. The existing framework uses a large global buffer, named \textit{arena}, and lets tensors share the memory space depending on their lifetimes. This work figures out that long-living tensors and large interim tensors for concatenation occupy lots of memory space. This work also finds that the existing tiny machine learning framework cannot effectively identify and handle the issues with the image segmentation models.

Based on the analysis, this work proposes TinySeg, a new model optimizing framework that enables memory-efficient image segmentation on tiny embedded systems. TinySeg analyzes the cold (idle) ranges of tensors in the target image segmentation model and identifies tensors that have long cold ranges. Then, TinySeg optimizes the peak memory usage of the target model mainly with two methods: (i) tensor spilling into local or remote storage to take cold tensors aside and (ii) fused fetching of spilled tensors to remove large interim tensors. At run-time, TinySeg efficiently implements tensor spilling and fetching through dynamic tensor compression and asynchronous block operation.

To demonstrate the effectiveness of the proposed framework, this work develops the TinySeg framework on top of the existing tiny machine learning framework, TensorFlow Lite for Microcontrollers, with new operators for tensor spilling and fused fetching. This work also designs a tiny image segmentation model to enable image segmentation on a small microcontroller board with a main memory of 1 megabyte. This work evaluates the TinySeg framework in terms of peak memory usage, network latency, and power consumption. The evaluation results show that TinySeg can reduce the peak memory usage of the image segmentation model by 39.3\% at most, enabling more intelligent image segmentation on low-power embedded systems.

The contributions of this work are:
{
\setlist[itemize]{leftmargin=7mm}
\begin{itemize}
    \item A new model optimizing framework, called TinySeg, for image segmentation on tiny embedded systems.
    \item Design of the TinySeg model optimizer and the TinySeg runtime with a variety of optimization methods.
    \item Implementation and evaluation of the TinySeg framework on top of an existing machine learning framework.
\end{itemize}
}
\section{Background \& Motivation}
\label{sec:moti}

\subsection{Image Segmentation Networks}

In recent years, the computer vision community has developed various neural network architectures for accurate image segmentation. Image segmentation is to classify every pixel of an input image into a certain category, corresponding to the object that the pixel belongs to. U-Net is one of the most popular image segmentation networks, which is originally designed for medical image segmentation~\cite{ronneberger:2015:miccai}. Many state-of-the-art image segmentation networks, including V-Net~\cite{milletari:2016:3dv} and Swin-Unet~\cite{cao:2023:eccv}, have similar architectural structures to U-Net.

U-Net is said to have an \textit{hourglass}-like structure as illustrated in Figure~\ref{fig:unet}. Going downwards, it repeatedly applies convolutions and poolings, and then activations become smaller in height and width but deeper with more channels. Then, going upwards, it repeatedly applies transposed convolutions and normal convolutions, and then activations recover the original resolution. Meanwhile, the early activations are concatenated with the later activations via long skip connections for the final prediction.

\subsection{TensorFlow Lite for Microcontrollers}

\begin{figure}[t]
    \centering
    \includegraphics[width=.9\hsize]{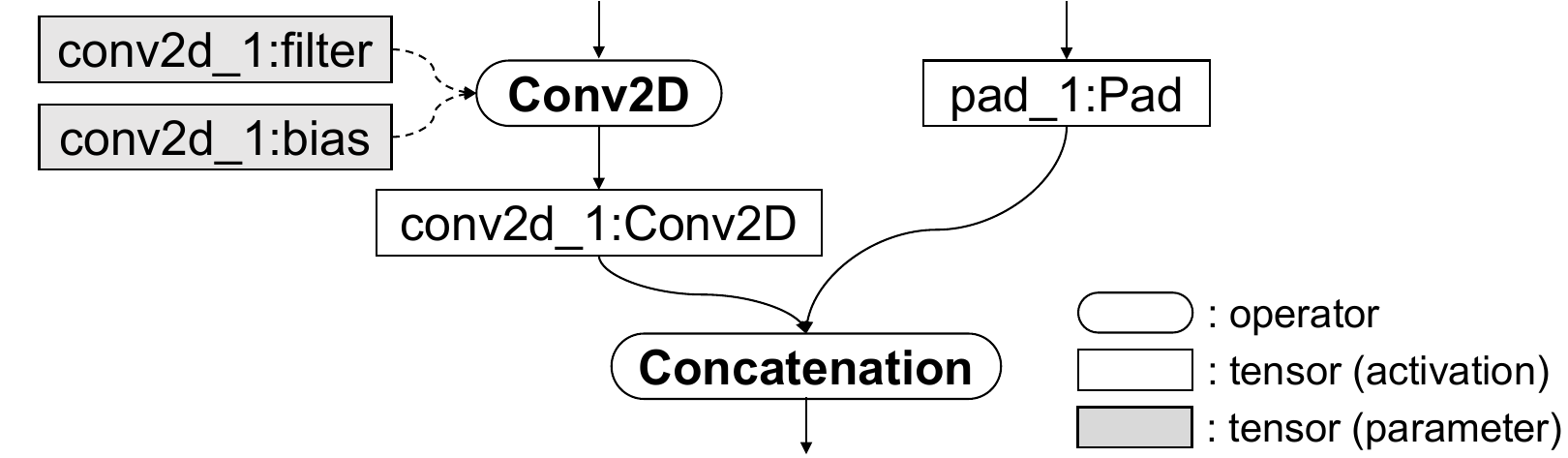}
    \caption{Neural Network Model in a Graph Representation.}
    \label{fig:graph}
\end{figure}

TensorFlow Lite for Microcontrollers (TFLM)~\cite{david:2021:tflm} is a tiny machine learning framework that facilitates deploying neural network models on low-power microcontrollers, such as Arduino microcontroller boards. TFLM provides a C++ programming interface to load and run a neural network model on bare-metal microcontrollers. TFLM represents a neural network model as a graph of operators and tensors, where each operator takes input tensors, performs calculations, and returns output tensors, like in Figure~\ref{fig:graph}.

As microcontrollers have limited memory of less than a megabyte, memory management is a crucial part of the framework. In general, neural network models need to have two types of data in memory:
\begin{itemize}
    \item \textbf{Parameters}: Some neural network operators include parameters (weights), which are used to perform calculations, such as convolution filters. For example, in Figure~\ref{fig:graph}, \texttt{conv2d\_1:filter} and \texttt{conv2d\_1:bias} are the parameters of \texttt{Conv2D}. The memory usage for parameters is usually \textit{static} because the parameter values do not change after training in general.
    \item \textbf{Activations}: During execution, neural network operators generate intermediate outputs (feature maps), which become the inputs of other operators. For example in Figure~\ref{fig:graph}, \texttt{conv2d\_1:Conv2D} is the activations defined by the \texttt{Conv2D} operator and used by the \texttt{Concatenation} operator. The memory usage for activations is usually \textit{dynamic} because it depends on how memory space is shared among tensors.
\end{itemize}

TFLM uses a fixed-size global memory buffer called \textit{arena} and stores the activations in the global buffer from when they are created to when they are last used (i.e., during their lifetimes). TFLM analyzes the lifetime of each tensor and generates a memory plan considering the lifetimes and sizes of tensors. TFLM allows tensors with non-overlapping lifetimes to share the same memory space so that it can minimize the peak memory usage for running a network model.

When loading the model, TFLM checks whether the peak memory usage of a model will be larger than the arena size specified by the user. If so, TFLM returns an out-of-memory error, and the model \textit{cannot} run on the target system. That is, depending on the dynamic memory usage, the model may fail to run on the target system.

\begin{figure}[t]
    \centering
    \begin{subfigure}{\hsize}
        \includegraphics[width=\hsize]{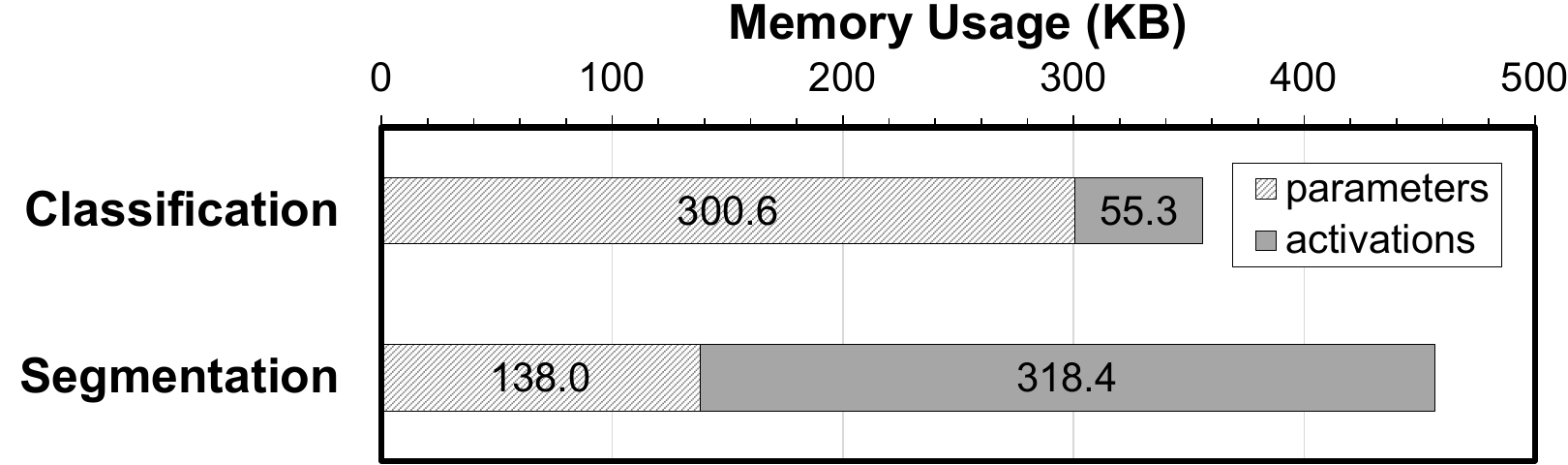}
        \caption{Memory Usage}
        \label{fig:memory:usage}
    \end{subfigure}\\\vspace{10pt}
    \begin{subfigure}{\hsize}
        \includegraphics[width=\hsize]{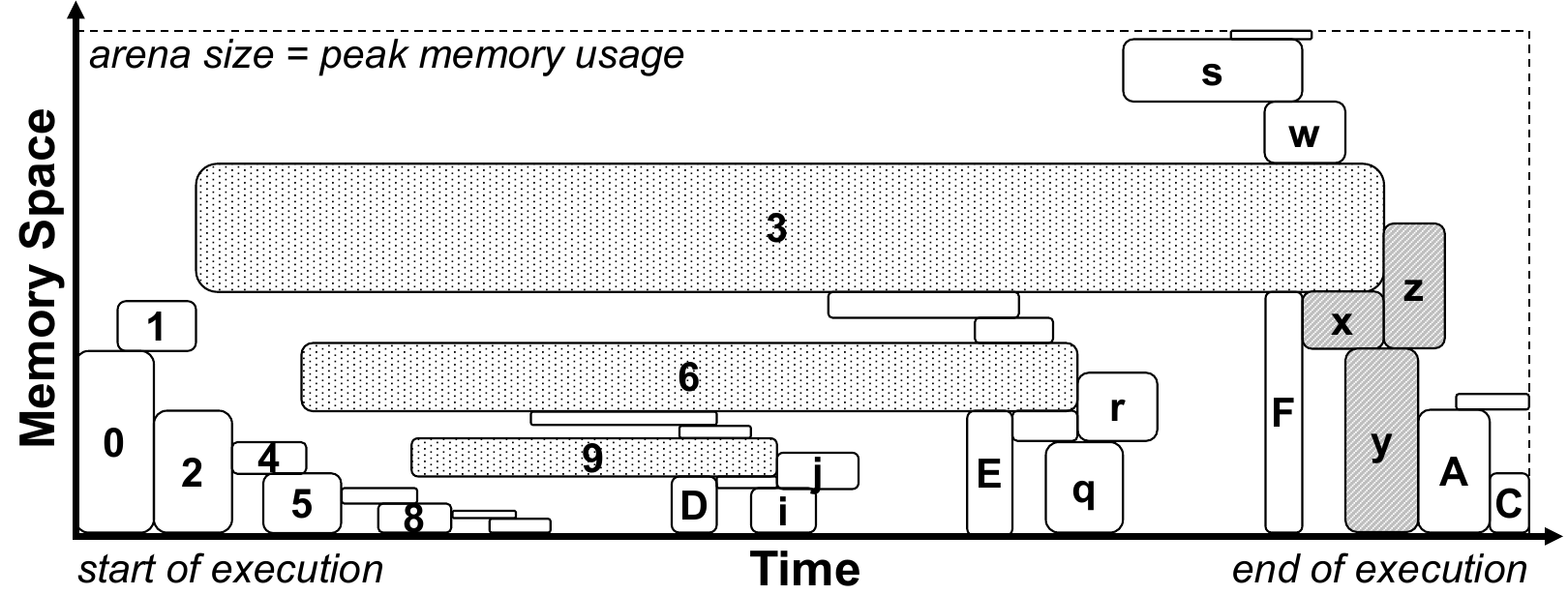}
        \caption{Memory Plan}
        \label{fig:memory:plan}
    \end{subfigure}
    \caption{Memory Usage of an Image Segmentation Model.}
    \label{fig:memory}
\end{figure}

\subsection{Memory Usage of Image Segmentation Models}

This work analyzes the memory usage of an image segmentation model on top of the TFLM framework and compares the model with a typical convolutional neural network model for image classification, provided by the framework~\cite{person_detect:web}.

Figure~\ref{fig:memory}(\subref{fig:memory:usage}) shows the memory usage of the classification and segmentation models for both their parameters and activations. The graph reveals that the segmentation model requires much more memory for activations than parameters, while the classification model requires much less memory for activations. Then, the segmentation model is 2.18$\times$ smaller than the classification model in binary, but it requires 5.76$\times$ more memory for execution.

This work finds that the high peak memory usage of the image segmentation model comes from two major issues: (i) \textit{long-living tensors} that remain idle for a long time in memory and (ii) \textit{large interim tensors} for concatenation that are used only for a short time.

First, this work figures that in image segmentation models, long-living tensors unnecessarily take up large memory space. As shown in Figure~\ref{fig:unet}, image segmentation networks have long skip connections to incorporate early (concrete) information for the final prediction. Accordingly, the early activations have to be stored in memory for a long time until they are concatenated with the later activations. That is, even though the early activations are not accessed for a long time, large memory space is reserved for the activations, increasing the peak memory usage of the models.

Figure~\ref{fig:memory}(\subref{fig:memory:plan}) illustrates the memory plan for an image segmentation model generated by the default memory planner of TFLM. The graph shows which part of memory space (i.e., arena) each tensor occupies over time. For example, tensor 0 occupies 115.2 kilobytes in the bottom of the arena from time 0 to time 1. The graph demonstrates that there exist large long-living tensors such as tensor 3 and tensor 6, occupying the memory space for a long time.

This work also discovers that large interim tensors, especially for concatenation, also contribute to high peak memory usage. In Figure~\ref{fig:memory}(\subref{fig:memory:plan}), a concatenation operator reads tensor 3 and tensor x and writes to tensor y. Then, a convolution operator reads tensor y and writes to tensor z. Here, the interim tensor y is used only for a short time just to concatenate tensor 3 and tensor x but its memory usage is not negligible.

\begin{figure}[t]
    \centering
    \includegraphics[width=\hsize]{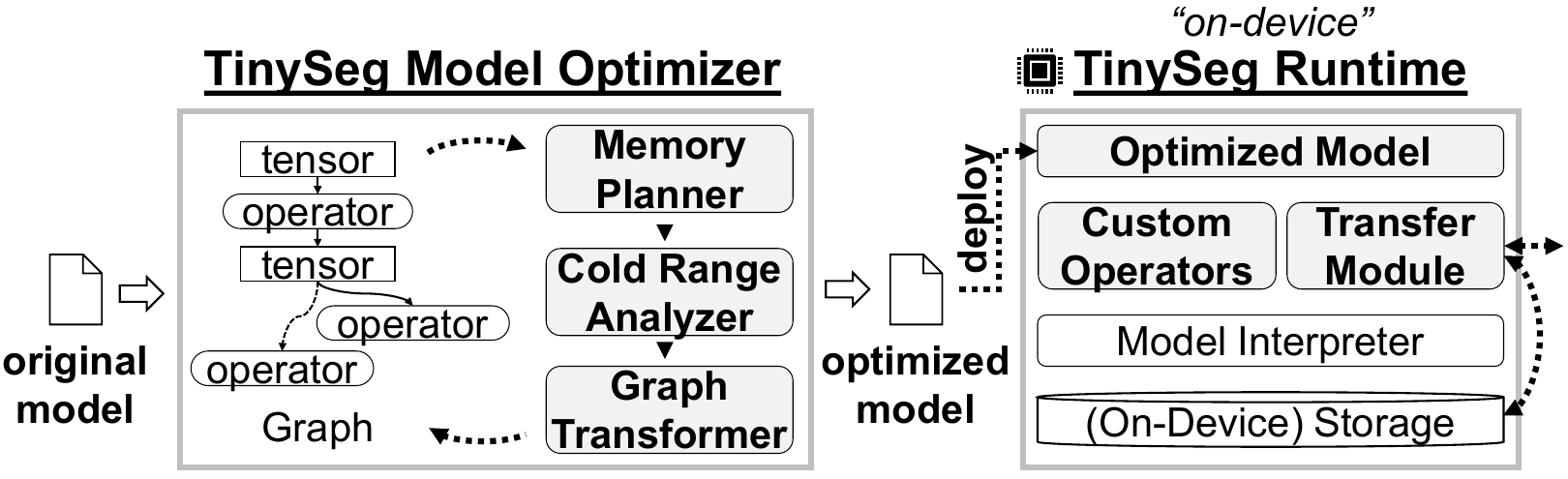}
    \caption{Overview of the TinySeg Optimizing Framework.}
    \label{fig:overview}
\end{figure}
\section{TinySeg: Model Optimizing Framework}

\begin{algorithm}[t]
\footnotesize
\caption{Model Optimization Process}
\label{alg:optimization}
\KwIn{Original graph $G$\\
\qquad\quad Target peak memory usage $M^*$}
\KwOut{Optimized graph $G_{opt}$}
$G_{opt} \leftarrow$ Copy the original graph $G$\\
$P_{opt}, M_{opt} \leftarrow $ Generate the memory plan of $G$ and calculate the peak memory usage of $G$\\
\While{$M_{opt}$ < $M^*$}{
    $CR_{opt} \leftarrow$ Perform cold range analysis w/ $G_{opt}$ and $P_{opt}$\\
    $G_{new} \leftarrow$ Transform $G_{opt}$ based on $CR_{opt}$\\
    $P_{new}, M_{new} \leftarrow $ Generate the memory plan of $G_{new}$ and calculate the peak memory usage of $G_{new}$\\
    \If{$M_{new} = M_{opt}$} {
        break
    }
    $G_{opt}, P_{opt}, M_{opt} \leftarrow$ $G_{new}, P_{new}, M_{new}$
}
\end{algorithm}

This work proposes TinySeg, an optimizing framework for memory-efficient image segmentation of which models are likely to have high peak memory usage. The TinySeg framework consists of the TinySeg model optimizer and the TinySeg runtime as illustrated in Figure~\ref{fig:overview}. The TinySeg model optimizer analyzes and transforms the input model to reduce its peak memory usage. The TinySeg runtime provides efficient custom operators to execute the optimized model.

The TinySeg model optimizer includes two main components for memory optimization: cold range analyzer and graph transformer. The cold range analyzer identifies the \textit{longest} cold range of each tensor in the input model where the tensor is stored in memory without being accessed. Next, the graph transformer modifies the model graph based on the analysis to handle long-living tensors and large interim tensors that contribute to peak memory usage.

The entire optimization process is summarized in Algorithm~\ref{alg:optimization}. At every iteration, the model optimizer analyzes the cold ranges of the tensors in the model, transforms the model based on the analysis, and generates a new memory plan to check the peak memory usage of the transformed model. This is because the memory plan of the model may change every time the graph transformer modifies the model and updates the lifetimes of tensors. If no further optimization is possible, the model optimizer stops optimizing the model and returns the optimized model.

Note that the TinySeg model optimizing framework imposes \textit{no restriction on} the type of the input model. Thus, the TinySeg framework can process any machine learning model even though it is most effective on image segmentation models that have long skip connections.

\begin{algorithm}[t]
\small
\caption{Cold Range Analysis}
\label{alg:analysis}
\KwIn{Graph $G$}
\KwOut{List of cold ranges $CR$}
\For{$op \in G.operators$}{
    \For{$t \in op.outputs$}{\label{alg:analysis:init:start}
        $CR[t].\{start, end, last\} \leftarrow op.id$ \\
    }\label{alg:analysis:init:end}
    \For{$t \in op.inputs$}{\label{alg:analysis:update:start}
        \If{$CR[t].end - CR[t].start < op.id - CR[t].last$} {
            $CR[t].start \leftarrow CR[t].last$ \\
            $CR[t].end \leftarrow op.id$ \\
        }
        $CR[t].last \leftarrow op.id$\\
    }\label{alg:analysis:update:end}
}
\end{algorithm}

\section{The TinySeg Model Optimizer}
\label{sec:optimizer}

\subsection{Cold Range Analyzer}

The cold range analyzer traverses the model graph from the root node and tracks where each tensor is defined and used to generate the cold range information. The cold range information for a tensor includes the following three integers:
\begin{itemize}
    \item \texttt{\textbf{start}}: the identifier of the operator at the start of the longest cold range
    \item \texttt{\textbf{end}}: the identifier of the operator at the end of the longest cold range
    \item \texttt{\textbf{last}}: the identifier of the operator that last uses the tensor
\end{itemize}
Note that the identifier of the \texttt{start} operator is always less than or equal to the identifier of the \texttt{end} operator.

Algorithm~\ref{alg:analysis} describes how the cold range analyzer builds the cold range information to be used in graph transformation. Basically, it iterates over the entire list of operators in the input graph and updates the (longest) cold ranges of the tensors. For each operator, it first checks the output tensors of the operator and initializes the cold range of each output tensor (from Line~\ref{alg:analysis:init:start} to Line~\ref{alg:analysis:init:end}). In the model graph, each tensor is defined only once, so it is safe to initialize the cold range information when it is defined by an operator.

Next, it checks the input tensors of the operator and updates the cold range of each input tensor (from Line~\ref{alg:analysis:update:start} to Line~\ref{alg:analysis:update:end}). If a tensor is an input of an operator, it means that the tensor is used by the operator. It also indicates the end of the current cold range and the start of the new cold range. Since the code range analyzer aims to find the \textit{longest} cold range, it first checks whether the current cold range ($op.id - CR[t].last$) is longer than the tentative longest cold range ($CR[t].end - CR[t].start$). If so, it updates the longest cold range and then the \texttt{last} operator.

\begin{figure}
    \centering
    \includegraphics[width=\hsize]{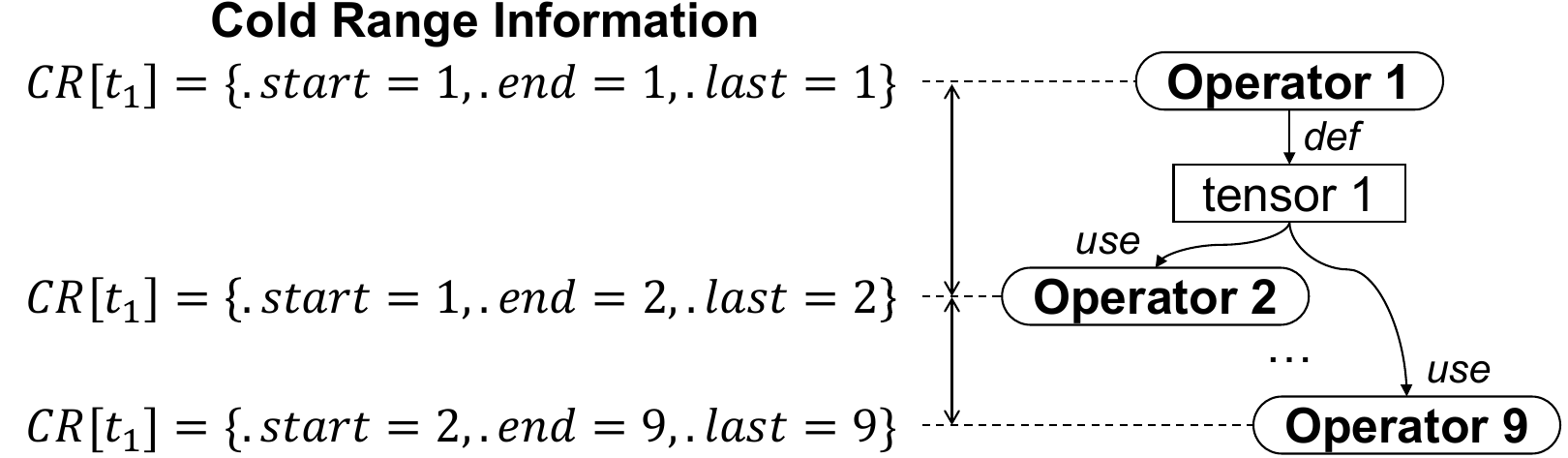}
    \caption{Example of Cold Range Analysis.}
    \label{fig:cold_range}
    \vspace{-10pt}
\end{figure}

Figure~\ref{fig:cold_range} illustrates an example cold range analysis for a single tensor. When \texttt{tensor 1} is defined by \texttt{Operator 1}, the analyzer initializes the cold range information of the tensor. Next, when the tensor is used for the first time, the analyzer updates the cold range information to commit the first cold range, which is from \texttt{Operator 1} to \texttt{Operator 2}. Then, when the tensor is used again by \texttt{Operator 9}, the analyzer compares the current cold range, which is from \texttt{Operator 2} to \texttt{Operator 9}, with the longest cold range. Since the current cold range is longer than the longest cold range, the analyzer updates the cold range information. In this way, the cold range analyzer builds the cold range information for all the tensors in the input model graph.

\subsection{Graph Transformer}

The graph transformer modifies the graph of the target model based on the cold range analysis with two methods: (i) tensor spilling into local or remote storage to take cold tensors aside and (ii) fused fetching to remove large interim tensors. Algorithm~\ref{alg:transform} briefly describes the overall transformation process.

\subsubsection{Tensor Spilling}

\begin{algorithm}[t]
\small
\caption{Graph Transformation}
\label{alg:transform}
\KwIn{Graph $G$\\
\qquad\quad List of cold ranges $CR$\\
\qquad\quad Target memory reduction $MR$}
\KwOut{Optimized graph $G_{opt}$}
\SetKwFunction{argmax}{argmax}
\SetKwFunction{spill}{Spill}
\SetKwFunction{fetching}{Fetching}
$G_{opt} \leftarrow$ Copy the original graph $G$\\
$ts \leftarrow $ Find tensors associated with peak memory usage \\\label{alg:transform:spill:start}
$victim \leftarrow ts[\argmax \{ CR[t].end - CR[t].start\ |\ t \in ts \}]$ \\\label{alg:transform:victim}
$op_{s} \leftarrow G_{opt}.operators[CR[victim].start]$ \\
$op_{e} \leftarrow G_{opt}.operators[CR[victim].end]$ \\
\If{the size of the victim tensor is larger than $MR$}{
    $op_{sp} \leftarrow$ Create a \texttt{Split} operator \\\label{alg:transform:partial:start}
    $op_{cc} \leftarrow$ Create a \texttt{Concatenation} operator \\
    Insert $op_{sp}$ after $op_{s}$ and $op_{cc}$ before $op_{e}$ to $G_{opt}$\\\label{alg:transform:partial:end}
    $victim \leftarrow op_{sp}.outputs.last$, $op_{s} \leftarrow op_{sp}$, $op_{e} \leftarrow op_{cc}$
}
$op_{spi} \leftarrow$ Create a \texttt{Spill} operator that spills $victim$ \\\label{alg:transform:insert:start}
$op_{fet} \leftarrow$ Create a \texttt{Fetching} operator that fetches $victim$ \\
Insert $op_{spi}$ after $op_{s}$ and $op_{fet}$ before $op_{e}$ to $G_{opt}$ \\\label{alg:transform:insert:end}
\label{alg:transform:spill:end}
\While{$op_{fet}$ is fusible with the following operator} {
    $op_{fet} \leftarrow $ Fuse $op_{fet}$ with the following operator
}
\end{algorithm}

To avoid unnecessary memory usage by long-living tensors, the model optimizer spills long-living tensors into local or remote storage if it is expected to be beneficial. For a given model graph, the graph transformer finds a tensor to spill and modifies the graph by spilling the tensor. To choose the victim tensor, it finds the tensors associated with peak memory usage. Among the tensors, the graph transformer selects the tensor with the longest cold range as the victim tensor (Line~\ref{alg:transform:victim} in Algorithm~\ref{alg:transform}).

After selecting the victim tensor, the optimizer checks if the size of the victim tensor is larger than the target amount of memory reduction. If so, it means that spilling the entire tensor is unnecessary to achieve the target memory reduction. To partially spill the victim tensor, the optimizer inserts additional operators that split the victim tensor into two sub-tensors and concatenate the sub-tensors (from Line~\ref{alg:transform:partial:start} to Line~\ref{alg:transform:partial:end} in Algorithm~\ref{alg:transform}). Then, one of the sub-tensors becomes the new victim tensor, and the start and end operators are updated accordingly. In this way, the optimizer can decrease the size of the tensor to be spilled, reducing the unnecessary overhead of tensor spilling and fetching.

\begin{table}[b]
    \small
    \centering
    \setlength{\tabcolsep}{3pt}
    \caption{Inputs and Outputs of the Custom Operators}
    \label{tab:operators}
    \vspace{-6pt}
    \begin{tabular}{cccl}
    \toprule
        \textbf{Operator} & \textbf{Type} & \textbf{Name} & \textbf{Description} \\
    \midrule
        \texttt{Spill} & Input & \texttt{victim} & Tensor to spill \\
            & Option & \texttt{id} & Identifier of the victim tensor \\
    \midrule
        \texttt{Fetching} & Input & \texttt{tensors} & List of tensors to concatenate \\
            & Option & \texttt{victim} & Identifier of the tensor to fetch \\
            & Option & \texttt{nth} & Position of the victim tensor \\
            & Option & \texttt{axis} & Axis along which to concatenate \\
            & Output & \texttt{output} & Concatenated tensor \\
    \bottomrule
    \end{tabular}
\end{table}

Next, the graph transformer inserts operators that implement tensor spilling to the start and end of the cold range (from Line~\ref{alg:transform:insert:start} to Line~\ref{alg:transform:insert:end} in Algorithm~\ref{alg:transform}). TinySeg provides two custom operators for tensor spilling: \texttt{Spill} and \texttt{Fetching} operators. The \texttt{Spill} operator transfers tensor data to local or remote storage. The \texttt{Fetching} operator fetches the tensor from local or remote storage and concatenates it with the input tensor(s) if any.

\begin{figure}[t]
    \centering
    \begin{subfigure}{.49\hsize}
        \includegraphics[width=\hsize]{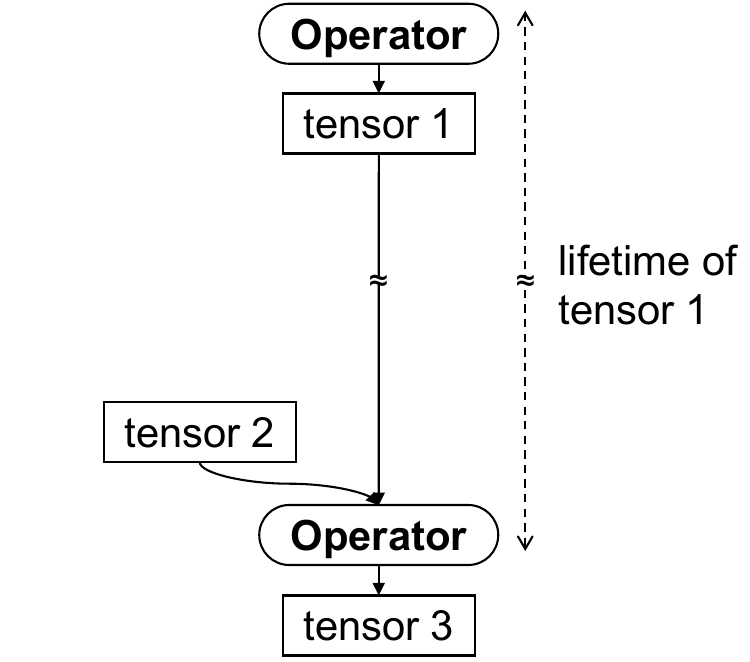}
        \caption{Before Spilling}
        \label{fig:spill:before}
    \end{subfigure}
    \begin{subfigure}{.49\hsize}
        \includegraphics[width=\hsize]{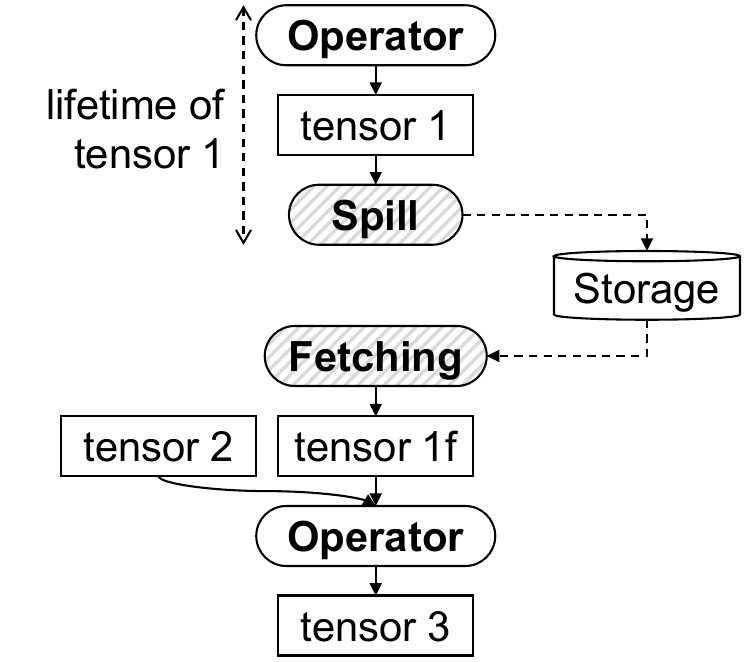}
        \caption{After Spilling}
        \label{fig:spill:after}
    \end{subfigure}
    \vspace{-5pt}
    \caption{Spilling a Tensor with a Long Cold Range.}
    \label{fig:spill}
    \vspace{-5pt}
\end{figure}

Figure~\ref{fig:spill} briefly illustrates how a model graph changes after tensor spilling. First, the graph transformer inserts the \texttt{Spill} operator at the start of the cold range. The \texttt{Spill} operator takes the victim tensor (e.g., tensor 1 in the figure) as its input and returns no output tensor. Next, the graph transformer inserts the \texttt{Fetching} operator at the end of the cold range. In the end, the lifetime of the victim tensor shortens because the tensor no longer has to be in memory during its cold range as shown in the figure. Then, the memory space occupied by the tensor can be shared with the other tensors.

\begin{figure}[t]
    \centering
    \begin{subfigure}{.49\hsize}
        \includegraphics[width=\hsize]{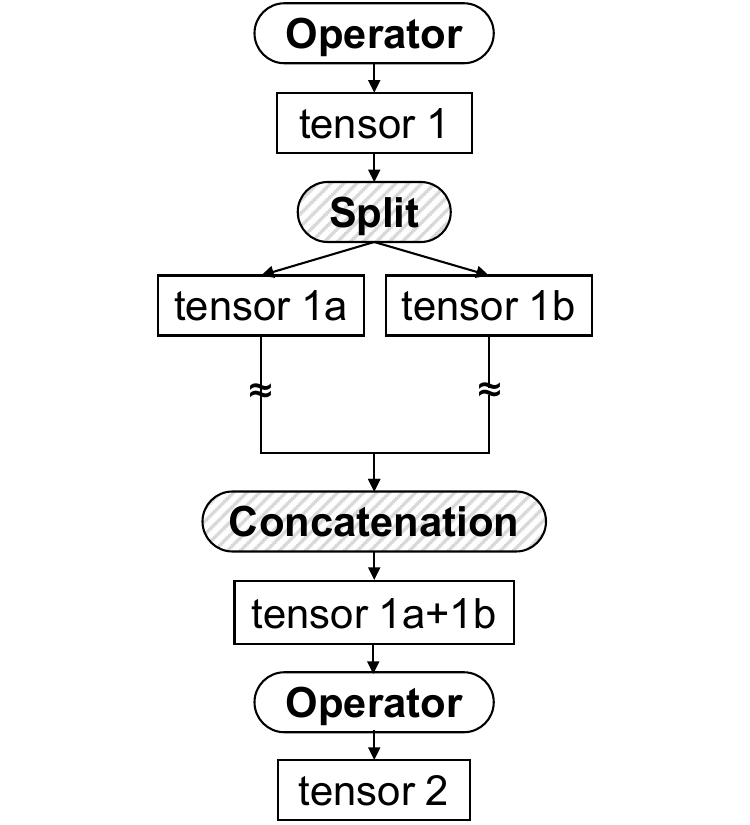}
        \caption{After Splitting}
        \label{fig:partial_spill:before}
    \end{subfigure}
    \begin{subfigure}{.49\hsize}
        \includegraphics[width=\hsize]{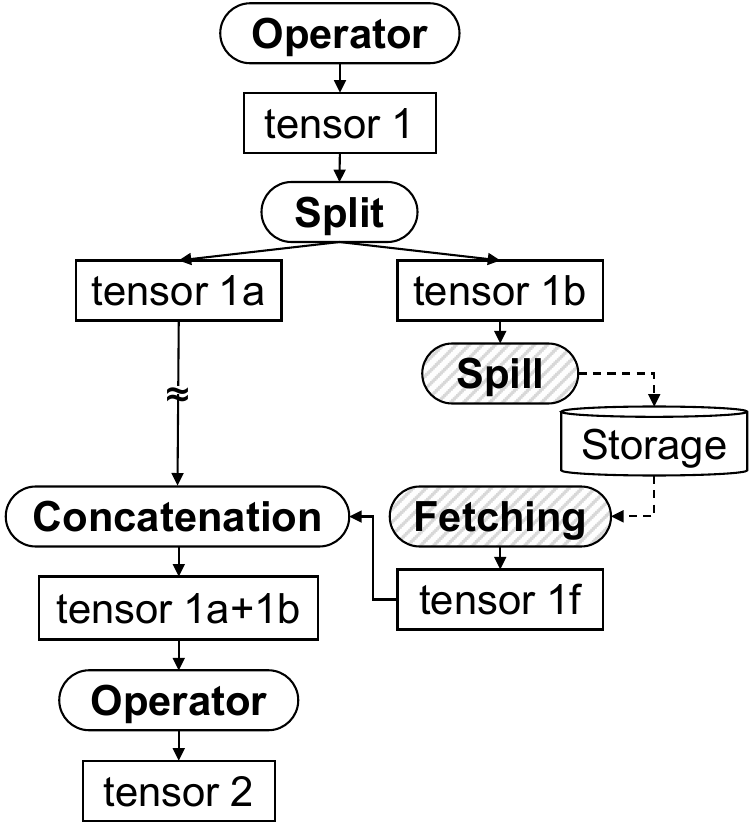}
        \caption{After Spilling}
        \label{fig:partial_spill:after}
    \end{subfigure}
    \vspace{-5pt}
    \caption{Partially Spilling a Tensor.}
    \label{fig:partial_spill}
    \vspace{-10pt}
\end{figure}

Figure~\ref{fig:partial_spill} describes how a model graph changes in the case of partial spilling. When partial spilling is allowed, the optimizer inserts a \texttt{Split} operator to divide the victim tensor into two sub-tensors. Then, the optimizer inserts a \texttt{Concatenation} operator to restore the original tensor with the two sub-tensors. Then, as one sub-tensor (i.e., \texttt{tensor 1b} in the figure) becomes the new victim tensor, the optimizer inserts the \texttt{Spill} and \texttt{Fetching} operators like in Figure~\ref{fig:spill}.

Note that the optimizer determines the split size considering the target amount of memory reduction. For example, assuming that the size of the victim tensor is $N$, the optimizer splits the victim tensor into one sub-tensor with the size of $N - \texttt{align}(MR)$ and another sub-tensor with the size of $\texttt{align}(MR)$, which becomes the new victim tensor. The optimizer applies the alignment according to the shape and data type of the victim tensor.

\subsubsection{Fused Fetching}

After spilling a tensor, the model optimizer tries to fuse the \texttt{Fetching} operator with the following operator if it is possible and beneficial. First, operator fusion may not be possible if the framework does not support an efficient fused operation. Second, operator fusion may not be beneficial if the interim tensor is smaller than the input and output tensors. If the interim tensor is smaller than the input and output tensors, the peak memory usage may increase after fusion. After the fusion, the fused operator would perform original calculations while fetching the spilled tensor from local or remote storage.

Especially, when the following operator is \texttt{Concatenation}, the optimizer merges the \texttt{Concatenation} operator into the \texttt{Fetching} operator. As the \texttt{Fetching} operator can internally perform concatenation, it is not necessary to have another \texttt{Concatenation} operator. Then, the \texttt{Fetching} and following \texttt{Concatenation} operators become a new \texttt{Fetching} operator after the fusion. For the other types of operators, the optimizer creates a new fused operator that performs two operations at the same time.

\begin{figure}[t]
    \centering
    \begin{subfigure}{.49\hsize}
        \includegraphics[width=\hsize]{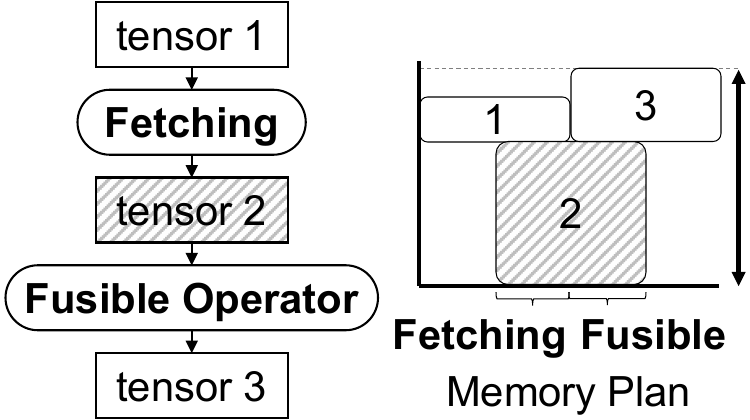}
        \caption{Before Fusion}
        \label{fig:fusion:before}
    \end{subfigure}
    \begin{subfigure}{.49\hsize}
        \includegraphics[width=\hsize]{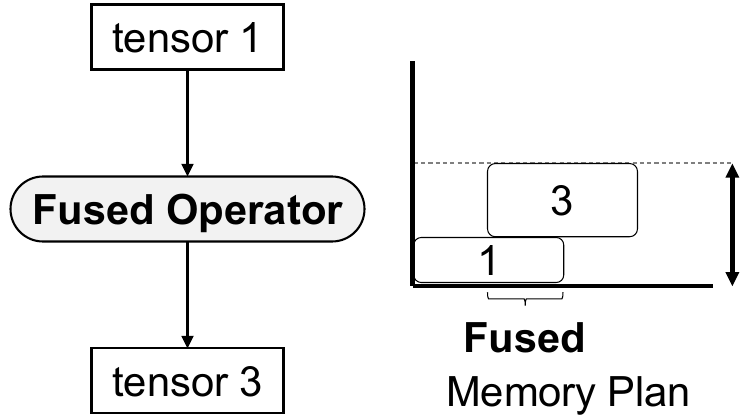}
        \caption{After Fusion}
        \label{fig:fusion:after}
    \end{subfigure}
    \caption{Fused Fetching After Spilling.}
    \label{fig:fusion}
\end{figure}

Figure~\ref{fig:fusion} briefly illustrates how a model graph changes after an operator fusion. The graph transformer fuses the two operators and removes the interim tensor because the tensor is no longer needed after fusion. Here, it is assumed that tensor 2 is larger than either tensor 1 or tensor 3. The peak memory usage before fusion is
\begin{align*}
max\{ | \text{\texttt{tensor 1}}| + |\text{\texttt{tensor 2}}|, |\text{\texttt{tensor 2}}| + |\text{\texttt{tensor 3}}| \}
\end{align*}
and then the peak memory usage becomes 
\begin{align*}
|\text{\texttt{tensor 1}}| + |\text{\texttt{tensor 3}}|
\end{align*} after operator fusion. Since tensor 2 is larger than either tensor 1 or tensor 3, the modified graph would have lower peak memory usage than the original graph. In this way, the memory optimizer can further optimize the memory usage of the target model.

\section{The TinySeg Runtime}
\label{sec:runtime}

The TinySeg runtime is in charge of implementing tensor spilling and fetching at run-time. The TinySeg runtime uses various optimization methods to reduce the performance overhead, which are described in this section.

\subsection{Dynamic Tensor Compression}

\begin{figure}[t]
    \centering
    \begin{subfigure}{.49\hsize}
        \includegraphics[width=\hsize]{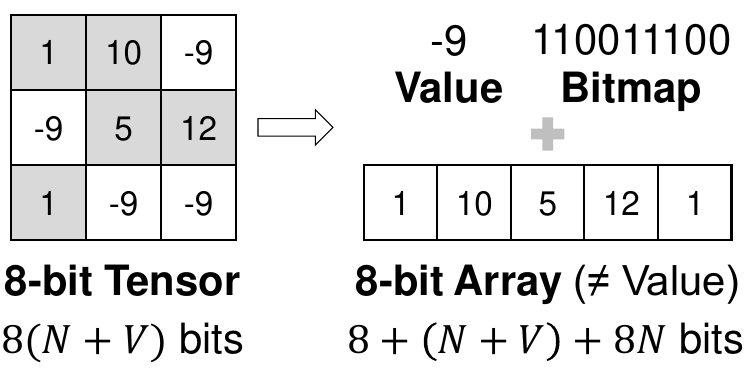}
        \caption{Method}
        \label{fig:run_comp:method}
    \end{subfigure}
    \begin{subfigure}{.49\hsize}
        \includegraphics[width=\hsize]{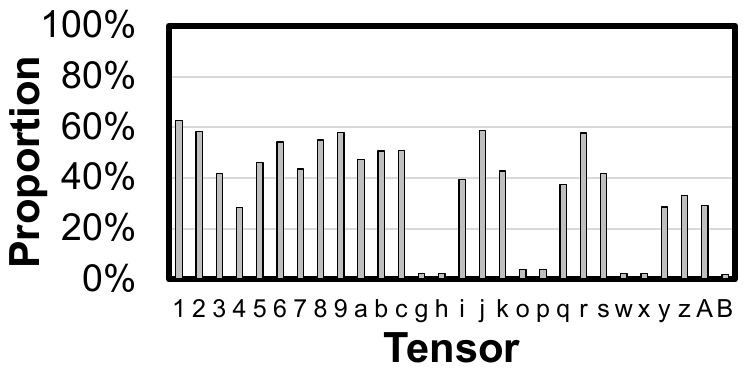}
        \caption{Distribution of Proportion}
        \label{fig:run_comp:dist}
    \end{subfigure}
    \caption{Dynamic Tensor Compression. ($V$: \# Occurrences of the Representative Value, $N$: \# Occurrences of the Other Values) }
    \label{fig:run_comp}
\end{figure}

The TinySeg runtime transfers tensor data for tensor spilling and fetching at run-time. To reduce data transfer overhead, this work designs a simple yet effective (compressed) tensor representation, which removes the duplicate occurrences of the most frequent value. As illustrated in Figure~\ref{fig:run_comp}(\subref{fig:run_comp:method}), it consists of a representative value, an array of the other values, and a bitmap to indicate the positions of the other values. For example in the figure, the most frequent value of the tensor is $-9$ and the bitmap indicates the position of each value that is not equal to the most frequent value. 

Although Compressed Sparse Row (CSR)~\cite{buluc:2009:spaa} is a widely-used representation for sparse tensors, it is more complicated and more suitable when the sparsity of tensors is high (i.e., when there are many zeros in the tensor data). However, this work finds that tensors in a sample image segmentation model often contain one very frequent non-zero value that occupies around 50\% of tensors. Figure~\ref{fig:run_comp}(\subref{fig:run_comp:dist}) shows the proportion of the most frequent value for each tensor in the sample image segmentation model. The graph shows that except for a couple of tensors, the representative values are dominant in the corresponding tensors. 

\begin{figure}[b]
    \centering
    \begin{subfigure}{\hsize}
        \includegraphics[width=\hsize]{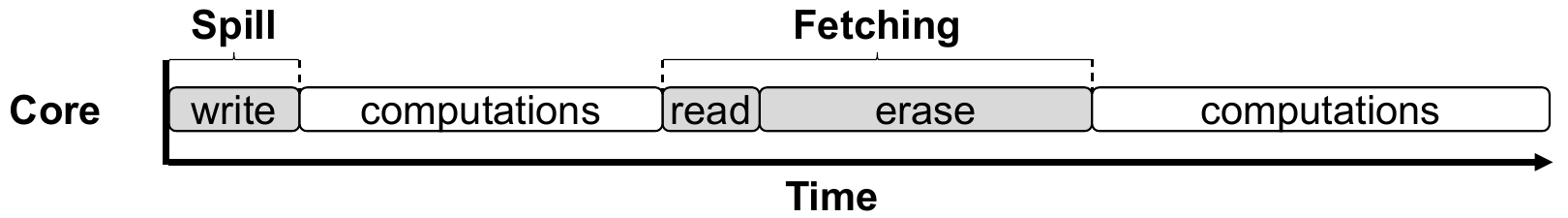}
        \caption{Synchronous}
        \label{fig:run_fetch:sync}
    \end{subfigure}\\
    \begin{subfigure}{\hsize}
        \includegraphics[width=\hsize]{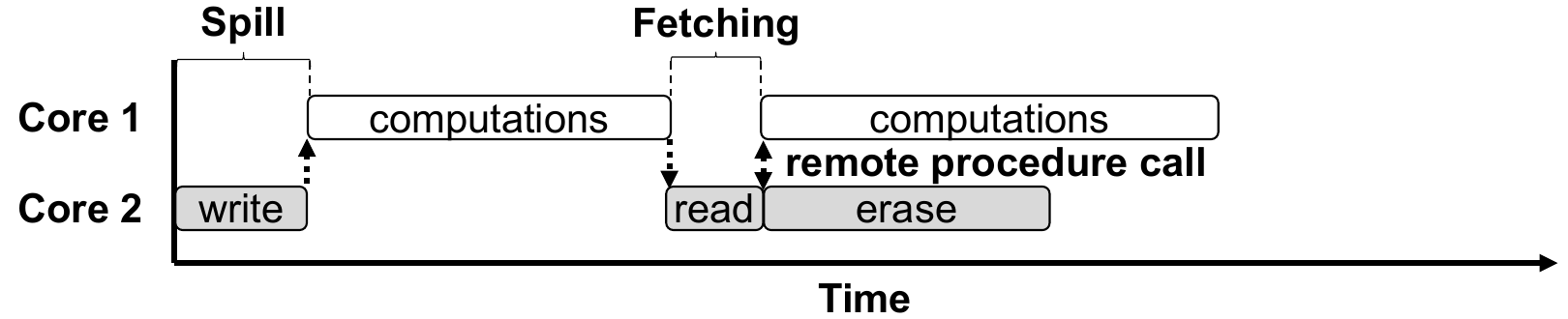}
        \caption{Asynchronous}
        \label{fig:run_fetch:asynch}
    \end{subfigure}
    \caption{Asynchronous Block Operation with Dual Cores.}
    \label{fig:run_fetch}
\end{figure}

\subsection{Asynchronous Block Operation}

Depending on the specification of the target storage, tensor spilling and fetching may incur non-negligible overhead, largely increasing the total network latency. Especially when using on-device flash memory, it could be more expensive because block erases are necessary before overwriting tensor data. In addition, block erase is known to take much longer than block write or read in typical flash memory.

To hide the block erase overhead, the TinySeg runtime enables asynchronous block operation using dual-core processing. Recently, many microcontrollers come with multiple processing cores. Since block erase has no dependence on the following computations within a single model execution, the runtime can use another core to asynchronously erase flash blocks. Therefore, as illustrated in Figure~\ref{fig:run_fetch}, the runtime invokes an asynchronous remote procedure call to another core so that block erase can run in parallel.

\subsection{Temporary Tensor Quantization}

On top of dynamic tensor compression and asynchronous block operation, this work finds another memory optimization opportunity in the existing framework~\cite{tflm:web}. For some operators, the framework allocates and uses temporary tensors as scratch buffers. However, in the case of certain operators like \texttt{TransposeConv}, temporary tensors could be even larger than its input and output tensors, increasing the peak memory usage. Thus, this work further quantizes large temporary tensors (from 32-bit to 16-bit) if possible.
\section{Evaluation}
\label{sec:eval}

\subsection{Experimental Setup}

This work implements the prototype of TinySeg on top of the open-source TensorFlow Lite framework~\cite{tflm:web}. This work develops the cold range analyzer and graph transformer conforming to the TensorFlow Lite model format for applicability. In addition, this work extends the framework with three new TensorFlow Lite operators to support tensor spilling and fused fetching: \texttt{Spill}, \texttt{Fetching}, and \texttt{FetchingConv2D}, considering the operators used in typical image segmentation models. The source code of the prototype implementation is available at \url{https://github.com/coslab-kr/tinyseg}.

This work also designs a new tiny image segmentation network, called Tiny U-Net to enable image segmentation on tiny embedded devices. The new network is a miniaturized version of U-Net~\cite{ronneberger:2015:miccai} with smaller convolution layers and fewer skip connections. To demonstrate its usefulness, this work trains the network from scratch and obtains a comparable validation accuracy  for an existing image segmentation dataset~\cite{carvana:web}. More specifically, the Tiny U-Net model obtained 87.7\% accuracy while the original U-Net with around 200$\times$ parameters obtained 92.9\% for the validation set. Table~\ref{tab:arch} summarizes the characteristics of the network.

\begin{table}[b]
\small
\centering
\caption{Tiny U-Net Characteristics ($N$: Batch Size)}
\label{tab:arch}
\begin{tabular}{ll}
\toprule
\textbf{Characteristic} & \textbf{Value} \\
\midrule
Input tensor shape $(N, H, W, C)$ & $(N, 80, 120, 3)$\\
Output tensor shape $(N, H, W, C)$ & $(N, 80, 120, 1)$\\
Number of model parameters & 109,725 \\
Binary size (w/ 8-bit integer quantization) & 138,040 bytes \\
Validation accuracy (w/ Carvana~\cite{carvana:web} dataset) & 87.7\% \\ 
\bottomrule
\end{tabular}
\end{table}

To show the effectiveness of the TinySeg framework, this work applies the optimization methods to the original Tiny U-Net model (with 8-bit integer quantization) step by step and evaluates the optimized models in terms of memory usage, network latency, and power consumption. To compare the power consumption of the original and optimized models, this work measures the electric current flowing into the target device using a Keysight U1232A digital multimeter of which the current measurement resolution is 0.01 $\mu$A.

\noindent For tensor spilling, this work investigates these three options:
\begin{itemize}
    \item spilling a tensor into the on-device \textit{internal} flash memory, on the microcontroller chip
    \item spilling a tensor into the on-device \textit{external} flash memory, connected to the microcontroller
    \item spilling a tensor into the \textit{remote} storage by transferring the tensor through serial communication
\end{itemize}
Note that using a different spilling option does not affect the peak memory usage of the model, but the latency of the model.

This work measures the peak memory usage of the original and optimized models with the default memory planner of the existing framework called \texttt{GreedyMemoryPlanner}~\cite{tflm:web}. In addition, this work measures the network latency of each model on a commercial low-power microcontroller board, Arduino Nicla Vision~\cite{arduino:web}, which uses the STM32H757AII6 microcontroller equipped with dual ARM Cortex-M7/M4 cores, 2 MB internal flash memory, and 1 MB main memory, and connected with 16 MB external QSPI flash memory.

\subsection{Results}

\subsubsection{Memory Usage}

\begin{figure}[t]
    \includegraphics[width=\hsize]{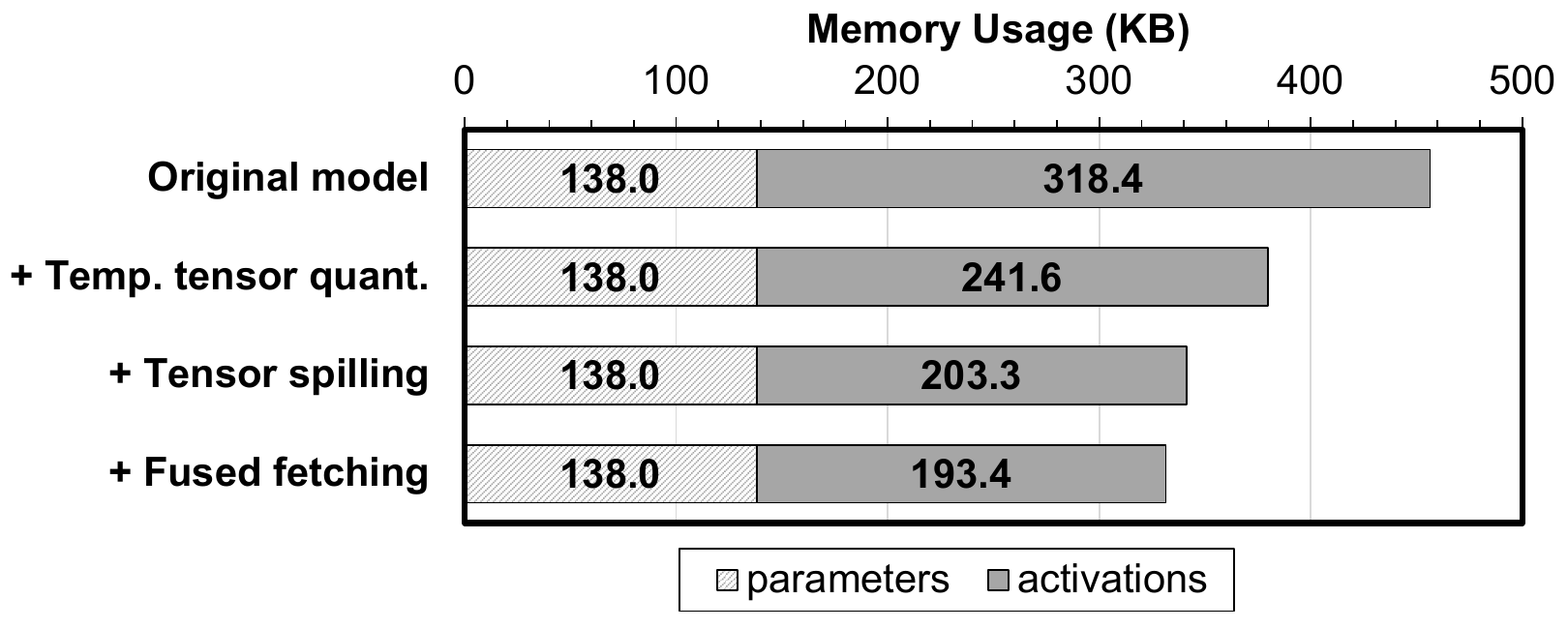}
    \vspace{-12pt}
    \caption{Total Memory Usage of the Models.}
    \label{fig:eval_usage}
\end{figure}

This work measures the peak memory usage of the original and optimized models with the prototype TinySeg optimizing framework. This work applies the three optimization methods to the original model in the following order: (1) temporary tensor quantization, (2) tensor spilling, and (3) fused fetching. During the optimization, the model optimizer selects one tensor (tensor 3 in Figure~\ref{fig:memory}(\subref{fig:memory:plan})) for tensor spilling as the optimizer finds that spilling another tensor will be no more beneficial. After tensor spilling, the model optimizer fuses two operators, \texttt{Fetching} and \texttt{Conv2D}, which then become \texttt{FetchingConv2D} after fusion.

Figure~\ref{fig:eval_usage} shows how the memory usage of the model changes when each optimization method is applied in the aforementioned order. Then, the bar at the bottom represents the memory usage of the final optimized model. As shown in the graph, the binary model size (i.e., the memory usage for parameters) changes little in each step. After tensor spilling or fused fetching, the optimized model will have new operators in its graph. However, the new operators require only a few parameters such as the tensor identifier for tensor spilling, so they hardly increase the model binary size.

Regarding dynamic memory usage for activations, which is more critical for execution, the graph shows that each optimization method can contribute to reducing the peak memory usage of the model. First, quantizing temporary tensors largely reduces the peak memory usage by 24.1\%, freeing some buffer space used by scratch buffers like tensor F in Figure~\ref{fig:memory}(\subref{fig:memory:plan}). Then, spilling a long-living tensor further reduces the peak memory usage by 15.9\% (36.2\% compared with the original model) by shortening the lifetime of the long-living tensor. In the end, the final optimized model has 39.3\% lower peak memory usage than the original model.

\begin{figure}[t]
    \includegraphics[width=\hsize]{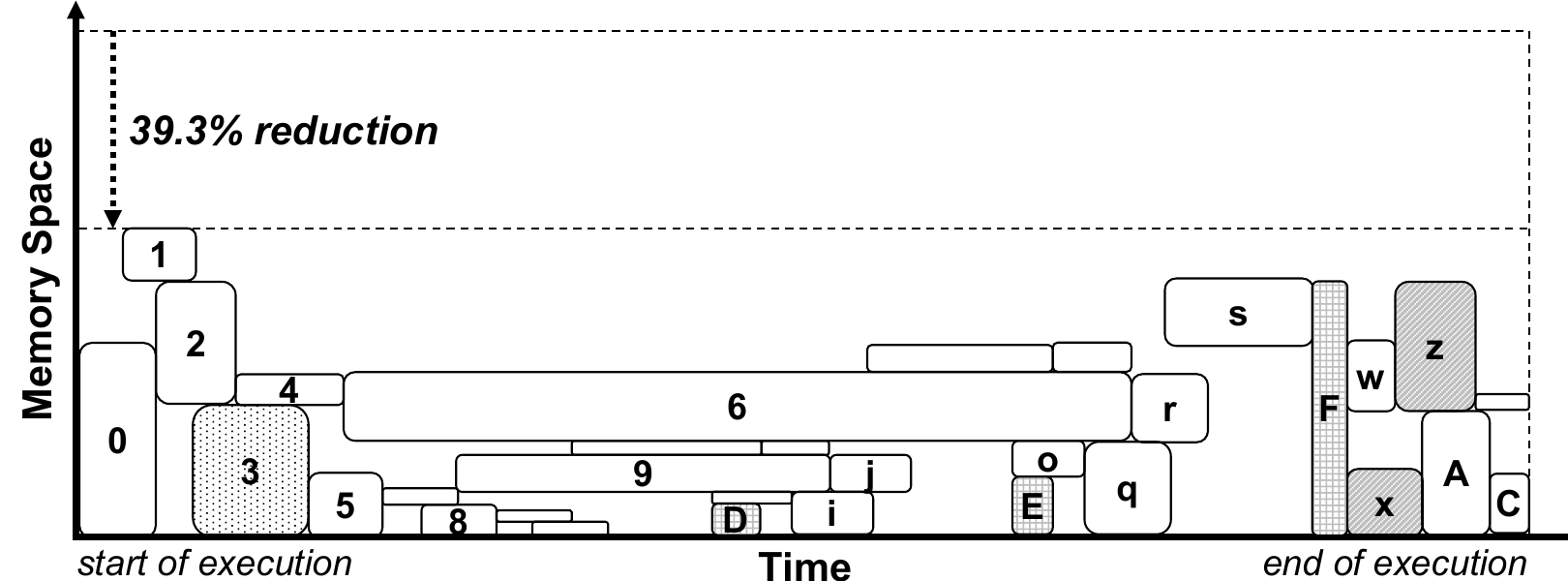}
    \vspace{-12pt}
    \caption{Memory Plan of the Final Optimized Model.}
    \label{fig:eval_plan}
\end{figure}

Figure~\ref{fig:eval_plan} illustrates the memory plan of the final optimized model, showing what changes in the memory plan result in the lower peak memory usage of the model, compared with the original one in Figure~\ref{fig:memory}(\subref{fig:memory:plan}). First, the figure shows that the lifetime of the long-living tensor (tensor 3) shortens so that the tensor can be placed in the low part of the memory space, leaving free memory space for other tensors. Second, the figure shows that the large interim tensor (tensor y between tensor x and tensor z in Figure~\ref{fig:memory}(\subref{fig:memory:plan})) is not present in the new memory plan because the interim tensor is no longer used by any operator with fused fetching. As a result, the new memory plan can better utilize the memory space than the original plan for image segmentation.

This work also compares the prototype framework with existing tiny machine learning frameworks for microcontrollers, as summarized in Table~\ref{tab:comparison}. In addition to TFLM, this work tries to optimize the same Tiny U-Net model with two recent frameworks: STM32Cube.AI~\cite{stm32cube.ai:web} and Pex~\cite{liberis:2023:pex}. Unfortunately, STM32Cube.AI fails to process the network raising an internal memory error due to a concatenation operator in the model. Although Pex is able to process the network, it generates exactly the same model after optimization, without enhancing the peak memory usage of the model.

\begin{table}[b]
\footnotesize
\centering
\caption{Comparison with Existing Tiny ML Frameworks}
\label{tab:comparison}
\begin{tabular}{llc}
\toprule
\textbf{Framework} & \textbf{Reference} & \textbf{Peak Memory Usage} \\
\midrule
TFLM~\cite{david:2021:tflm, tflm:web} (Baseline) & Commit 384dd27 & 318.4 KB \\
STM32Cube.AI~\cite{stm32cube.ai:web} & Version 8.1.0 & \textit{Internal memory error} \\
Pex~\cite{liberis:2023:pex, pex:web} & Commit 08709f1 & 318.4 KB \\
\textbf{This work} & - & \textbf{193.4 KB} \\
\bottomrule
\end{tabular}
\end{table}

Furthermore, this work applies the optimizations to large network models even though they do not fit in tiny embedded systems. This work selects three image segmentation models that can be converted to TensorFlow Lite models. Table~\ref{tab:large} summarizes the peak memory usages of the original and optimized models. The results show that the framework can successfully optimize the memory usages of other image segmentation models that have long skip connections.

\begin{table}[b]
\footnotesize
\centering
\caption{Peak Memory Usage of Large Network Models}
\label{tab:large}
\begin{tabular}{lccc}
\toprule
\textbf{Model} & \textbf{Model Size} & \textbf{Original} & \textbf{Optimized} \\
\midrule
V-Net (2D)~\cite{milletari:2016:3dv} & 4.30 MB & 1.35 MB & 1.09 MB \\
Attention U-Net~\cite{oktay:2018:midl} & 10.27 MB & 7.38 MB  & 3.19 MB \\
ResUnet-a~\cite{diakogiannis:2020:resunet-a} & 53.32 MB & 21.16 MB & 12.78 MB \\
\bottomrule
\end{tabular}
\end{table}

\subsubsection{Network Latency}

This work measures the total end-to-end network latency of the original and optimized models on the commercial low-power microcontroller board with internal, external, and remote spilling options:
\begin{itemize}
    \item \textbf{Internal Spilling}: The runtime stores the victim tensor in the internal flash memory of the microcontroller. Note that when using the both processing cores, the 2 MB internal flash is partitioned for the cores.
    \item \textbf{External Spilling}: The runtime uses the external 16 MB flash memory of the board, which is connected to the microcontroller and shared by the two cores.
    \item \textbf{Remote Spilling}: The runtime transfers the victim tensor to the host and retrieves the tensor from the host when needed.
\end{itemize}

Table~\ref{tab:latency} summarizes the latency of each model with internal, external, or remote spilling. After quantizing temporary tensors, the network latency actually slightly increases, unexpectedly. This work presumes the reason for the increase is that memory operations are more optimized for 32-bit operations than for 16-bit operations in the system. After applying tensor spilling, the latency increases due to the data transfer overhead. With dynamic tensor compression, the latency decreases because it reduces the amount of data to transfer with little compression and decompression overhead. Finally, the overhead of internal or external spilling is largely reduced by asynchronous block operation.

Regarding the different spilling options, internal spilling achieves the best latency among the three options at the end. Before applying synchronous block operation, internal spilling results in doubling the total latency of the model. This work finds that, especially for the internal flash memory, block write takes much less time than block erase. On the other hand, for the external flash memory, block write is almost as slow as block erase. Then, by hiding the overhead of block erase, internal spilling can obtain the best performance among the different spilling options.

\begin{table}[t]
\small
\centering
\caption{Total Network Latency}
\label{tab:latency}
\begin{tabular}{lccc}
\toprule
\multirow[c]{2}{*}{\textbf{Model}} & \multicolumn{2}{c}{\textbf{Latency}} \\
 & Internal & External & Remote \\
\midrule
Original model & \multicolumn{3}{c}{1190 ms} \\
+ Temp. tensor quant. & \multicolumn{3}{c}{1199 ms} \\
\multirow[t]{2}{*}{+ Tensor spilling} & 2421 ms & 2494 ms & 1583 ms\\
\multirow[t]{2}{*}{+ Fused fetching} & 2430 ms & 2511 ms & 1587 ms\\
\multirow[t]{2}{*}{+ Tensor compression} & 2333 ms & 2046 ms & 1470 ms\\
\multirow[t]{2}{*}{+ Async. block operation} & 1270 ms & 1718 ms & N/A \\
\bottomrule
\end{tabular}
\end{table}

\begin{figure}[t]
\includegraphics[width=\hsize]{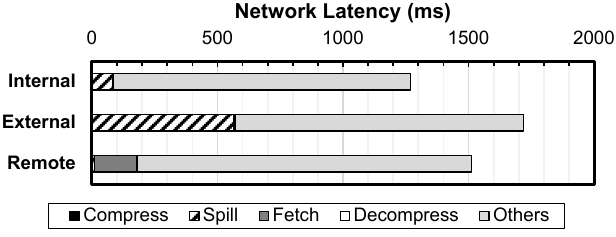}
\captionof{figure}{Breakdown of the Network Latency.}
\vspace{-5pt}
\label{fig:breakdown}
\end{figure}

Figure~\ref{fig:breakdown} shows the breakdown of the latency of the final model for each spilling option. The graph demonstrates that tensor spilling takes much more time with external spilling as block write is much slower with the external flash memory than the internal flash memory. For remote spilling, tensor spilling takes much less time than tensor fetching because the tensor data can be asynchronously buffered in the host for tensor spilling. In addition, the graph shows that tensor compression and decompression incurs almost no overhead compared to the total network latency.

\begin{figure}[t]
    \centering
    \begin{subfigure}{.49\hsize}
        \includegraphics[width=\hsize]{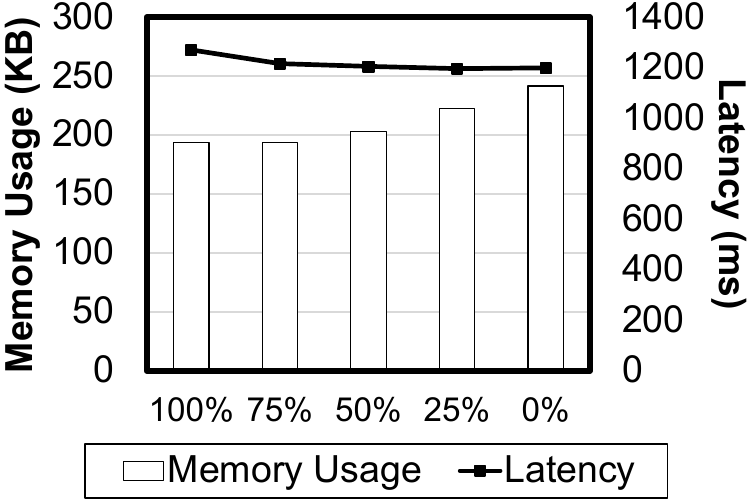}
        \caption{Internal Spilling}
        \label{fig:eval_partial:local}
    \end{subfigure}
    \begin{subfigure}{.49\hsize}
        \includegraphics[width=\hsize]{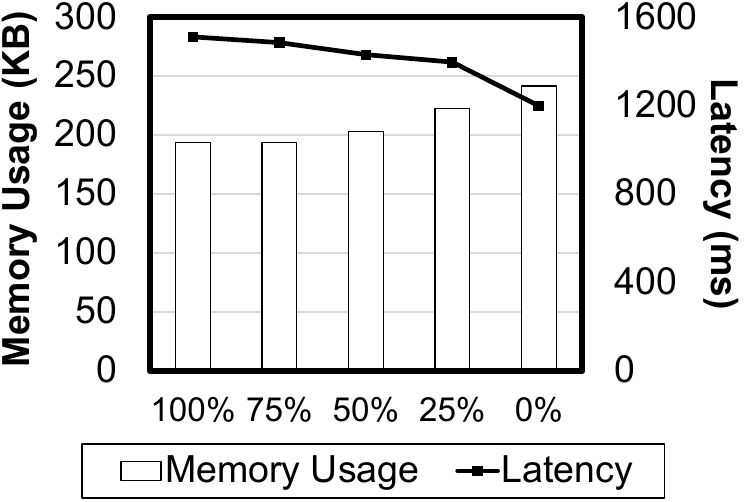}
        \caption{Remote Spilling}
        \label{fig:eval_partial:remote}
    \end{subfigure}
    \caption{Peak Memory Usage and Network Latency of the Optimized Model with Partial Spilling. }
    \label{fig:eval_partial}
\end{figure}

This work also measures how the memory usage and latency change with partial spilling. As explained in Section~\ref{sec:optimizer}, if the memory requirement is met, it is not necessary to spill the entire tensor, which would increase the runtime overhead. Figure~\ref{fig:eval_partial} shows the memory usage and network latency when spilling a different percentage of the victim tensor. 100\% spilling means spilling the entire tensor, and 0\% spilling means no spilling. As the percentage of spilling decreases, the peak memory usage increases but the network latency decreases due to the reduced transfer overhead. Interestingly, the peak memory usage increases slowly when the percentage of spilling is high. That is, partial spilling could be more cost effective than entire spilling when the memory requirement is not very strict.

In summary, the evaluation results demonstrate that the TinySeg framework can successfully reduce the memory requirement of an image segmentation model, enabling tiny embedded systems to implement smarter image segmentation. Although the current implementation incurs some performance overhead for data transfer, the network latency can be further optimized by using a faster storage or communication interface.

\subsubsection{Power Consumption}

\begin{figure}[b]
    \centering
    \begin{subfigure}{\hsize}
        \includegraphics[width=\hsize]{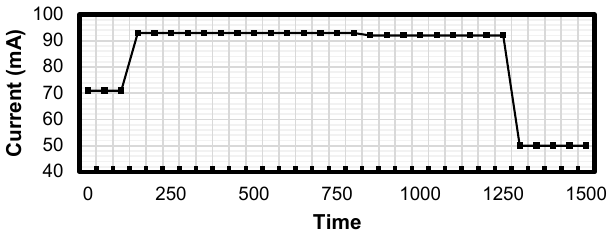}
        \caption{Original Model}
        \label{fig:eval_power:original}
    \end{subfigure}\\
    \begin{subfigure}{\hsize}
        \includegraphics[width=\hsize]{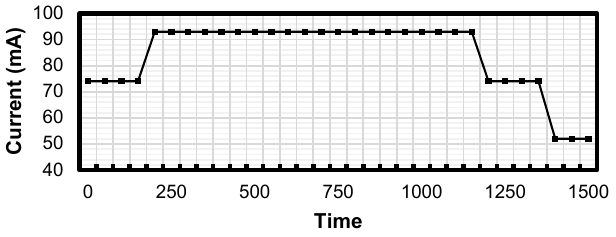}
        \caption{Optimized Model}
        \label{fig:eval_power:optimized}
    \end{subfigure}
    \caption{Current Flowing During Model Execution.}
    \label{fig:eval_power}
\end{figure}

This work evaluates the power consumption of the original and optimized models by measuring the current flowing into the microcontroller board. Figure~\ref{fig:eval_power} shows how the amount of current changes over time when executing the original and optimized models. For the optimized model, the internal flash memory is used for tensor spilling. On average, the board consumes less power when executing the original model than executing the optimized model. The average power consumption of the original model is 418.2 mW, while the average power consumption of the optimized model is 420.6 mW in the experiments.

The main reason for the increase is the use of another processing core. As explained in Section~\ref{sec:runtime}, the runtime uses another core to perform asynchronous block erase. When executing the original model, the runtime uses only one main core by default. However, the processing cores are heterogeneous (i.e., one Cortex-M7 core at up to 480 MHz and one Cortex-M4 core at up to 240 MHz), using the Cortex-M4 core does not increase the power consumption a lot.

\section{Related Work}
\label{sec:related}

\subsection{Framework-level Memory Optimization}

Previous work has proposed various framework-level methods for memory-efficient model execution on embedded systems~\cite{liberis:2023:pex, stm32cube.ai:web, lin:2021:mcunetv2, lee:2023:dac, ji:2021:jsa}.

Similar to this work, some recent work considers tiny embedded systems built with microcontrollers~\cite{liberis:2023:pex, stm32cube.ai:web, lin:2021:mcunetv2}. Liberis and Lane~\cite{liberis:2023:pex} introduce Pex, a partial execution compiler for memory-efficient deep learning on microcontrollers. Pex automatically identifies operators whose execution can be split along the channel dimension and generates memory-efficient execution schedules considering the operators. In addition, Pex applies structured pruning to the network model to further optimize the memory usage of the model.

STM32Cube.AI~\cite{stm32cube.ai:web} is a free tool from STMicroelectronics, which facilitates the optimization and deployment of neural network models for STM32 microcontrollers. It includes a memory optimizer for neural network models, which provides the visualization of model memory usage, the separation of model parameters for multiple storage, the reuse of input or output buffers for activations. It also includes a graph optimizer that supports various graph-level optimizations such as operator fusion and constant folding. 

Lin et al.~\cite{lin:2021:mcunetv2} propose an optimizing framework for executing neural network models on microcontrollers with limited memory. It reduces the peak memory usage of a model by processing the model patch by patch rather than processing the entire feature maps at a time. To efficiently implement the patch-by-patch execution, they introduce receptive field
redistribution, which reduces the receptive field of the initial stage. It also optimizes memory consumption by jointly designing the system and the network through neural architecture search.

Other work focuses on relatively larger embedded systems such as Nvidia Jetson platforms, which can run neural network models on top of operating systems~\cite{lee:2023:dac, ji:2021:jsa}. Lee et al.~\cite{lee:2023:dac} focus on reducing the memory burden on the embedded GPU during deep neural network inference. In the paper, they propose Occamy, a deep learning compiler that reduces the memory usage of network models without sacrificing accuracy by tensor coalescing, layer fusion, and memory access pattern analysis. It analyzes the liveness of tensors for layer fusion, merges input and output tensors to remove redundant tensors, and places tensors into memory pools to reduce memory management overhead.

Ji et al.~\cite{ji:2021:jsa} propose memory management schemes for model inference on edge-end embedded systems. First, they propose to load parameters in an incremental manner to reuse the memory space, enabling larger network models. Second, addressing the unnecessary memory consumption of blob and workspace tensors, they propose to reorganize the data layout of tensors in the middle of execution to better utilize the memory space.

Although the previous work demonstrates that their proposed methods can reduce the memory consumption of a given model for embedded systems, they do not consider long-living tensors and thus they cannot avoid storing the tensors in memory during their lifetimes. Therefore, their methods are relatively less effective for hourglass-like image segmentation models, while TinySeg is specially designed for image segmentation models with distinct architectural characteristics so that it can successfully optimize the models for better memory utilization.

\subsection{Model Compression Techniques}

Model compression is being widely studied in the artificial intelligence community, which is to reduce the memory consumption of a network model, generally for resource-constrained devices~\cite{choi:2017:iclr, gupta:2015:icml, jacob:2018:cvpr, banner:2019:nips, han:2015:nips, li:2017:iclr}. Especially, post-training model compression is to compress neural network models without retraining, thus easily applicable to existing well-trained models~\cite{jacob:2018:cvpr, banner:2019:nips, lazarevich:2021:iccvw, shi:2023:iccv}.

\textbf{Post-Training Quantization: } Jacob et al.~\cite{jacob:2018:cvpr} introduce an integer quantization scheme that quantizes both activations and weights as 8 bits with two quantization parameters (\textit{scale} and \textit{zero point}) and integer-only matrix multiplication based on the quantization scheme. The quantization scheme is the one used in the TensorFlow Lite framework. Banner et al.~\cite{banner:2019:nips} propose a practical 4-bit post-training quantization scheme. It reduces the accuracy loss from quantization using analytical clipping, per-channel bit allocation, and etc.

\textbf{Post-Training Pruning: } Lazarevich et al.~\cite{lazarevich:2021:iccvw} propose a method for post-training weight pruning that can obtain high sparsity rate but little accuracy drop. It selects layer-wise sparsity rates to achieve the global sparsity rate. More recently, Shi et al.~\cite{shi:2023:iccv} integrate pruning and quantization and optimize them  considering the interaction between them under the post-training setting.

Since the optimization methods of TinySeg are orthogonal to the model compression techniques, the TinySeg optimizing framework can be applicable in combination with the techniques, further optimizing the target network model for tiny embedded systems.
\section{Conclusion}
\label{sec:conclusion}

This work proposes TinySeg, a new model optimizing framework that enables memory-efficient image segmentation on tiny embedded systems. TinySeg analyzes the lifetimes of tensors in the target model and identifies tensors idle for a long time. Then, TinySeg optimizes the peak memory usage of the target model with two methods: (i) tensor spilling into local or remote storage and (ii) fused fetching of spilled tensors. This work implements TinySeg on top of the existing tiny machine learning framework and demonstrates that TinySeg can reduce the peak memory usage of an image segmentation model by 39.3\% for tiny embedded systems.

\balance
\bibliographystyle{ACM-Reference-Format}
\bibliography{paper}

\appendix









\end{document}